\pdfoutput=1

\documentclass[11pt]{article}

\usepackage[]{emnlp2021}

\usepackage{times}
\usepackage{latexsym}
\usepackage[bottom]{footmisc}

\usepackage[T1]{fontenc}

\usepackage[utf8]{inputenc}

\usepackage{microtype}
\usepackage{multirow}
\usepackage{wrapfig}
\usepackage{natbib}
\usepackage{color}
\usepackage{graphicx,enumitem}
\usepackage{colordvi}
\usepackage{subfigure}
\usepackage{amssymb}
\usepackage{amsmath}
\usepackage{amsthm}
\usepackage{hyperref}       
\usepackage{url}            
\usepackage{booktabs}       
\usepackage{amsfonts}       
\usepackage{nicefrac}       
\usepackage{microtype}  
\title{Regularized Training of Nearest Neighbor Language Models}

\author{Jean-Fran\c cois Ton* \\
  University of Oxford \\
  \texttt{ton@stats.ox.ac.uk} \\\And
  Walter Talbott \\
  Apple\And
  Shuangfei Zhai \\
  Apple\And
  Josh Susskind \\
  Apple}

\begin{document}
\maketitle
\begin{abstract}
 Including memory banks in a natural language processing architecture increases model capacity by equipping it with additional data at inference time. In this paper, we build upon $k$NN-LM \citep{khandelwal20generalization}, which uses a pre-trained language model together with an exhaustive $k$NN search through the training data (memory bank) to achieve state-of-the-art results. We investigate whether we can improve the $k$NN-LM performance by instead training a LM with the knowledge that we will be using a $k$NN post-hoc. We achieved significant improvement using our method on language modeling tasks on \texttt{WIKI-2} and \texttt{WIKI-103}. The main phenomenon that we encounter is that adding a simple L2 regularization on the activations (not weights) of the model, a transformer, improves the post-hoc $k$NN classification performance. We explore some possible reasons for this improvement.  In particular, we find that the added L2 regularization seems to improve the performance for high-frequency words without deteriorating the performance for low frequency ones.
\let\thefootnote\relax\footnotetext{$^*$\text{Work done as intern at Apple}}

\end{abstract}

\section{Introduction}
The problem of language modelling (LM) usually consists of two main challenges. Firstly, mapping the context, i.e. the sentence prefixes, to a vector representation, and secondly using this representation to predict the subsequent word. In \citep{khandelwal20generalization}, the authors claim that the first problem is much easier to solve. Hence, given a pre-trained LM, they post-hoc modify the representation using a \textit{k}-nearest neighbor scheme ($k$NN) and achieve significant improvements on challenging datasets, such as \texttt{WIKI-103}.

Given that $k$NN improves the overall language modelling of a pre-trained network, we examine training strategies that can make the underlying network's representations more amenable to the $k$NN step.  Our results show improvements over applying $k$NN to a generic LM network.

We first explore a simple learning scheme for the language model, where during training we intentionally push representations that predict the same word closer to together in the L2 sense, using a Momentum Contrastive (MOCO) \cite{he2020momentum} style implementation. We go on to note that this MOCO style learning can be replaced by simply adding L2 regularization to the activation of the layer used for $k$NN, eliminating implementation complexity.  We then present some initial experiments toward understanding why this L2 regularization brings improved performance.    

\section{Background}

Our work builds upon $k$NN-LM \citep{khandelwal20generalization}. In essence, $k$NN-LM tackles the problem of how to improve a trained LM's representations, and how to adapt LMs to capture non-frequent sentences that are usually forgotten by the model during training. $k$NN-LMs achieve significantly higher performance through a simple interpolation between the original LM predictions and the $k$NN predictions.

At inference time, given a new context sentence, $k$NN-LM works as follows:
\begin{enumerate}
    \item The context sentence $c_i$ is passed through the pre-trained network to produce a representation $r^{context}_{i} \in \mathbf{R}^d$ as well as the corresponding logits $y_i^{LM}$ to predict the next word.
    \item $r^{context}_i$ is used to find the $k$-nearest neighbors in the training data. The logits $y^{kNN}$ are computed by a weighted average of the neighbors' labels, using the inverse exponential distance as the weight for each neighbor.
    \item The logits are interpolated to give the final prediction: 
    $$
    y_{final} = \lambda y^{kNN} + (1-\lambda) y^{LM},
    $$
    where $\lambda$ is the interpolation parameter that can be tuned on validation data.
\end{enumerate}
This simple post-hoc implementation allows \citep{khandelwal20generalization} to improve upon the SOTA in LM by a significant margin. One thing to note about $k$NN-LM is that they do not need to retrain the LM and hence the whole algorithm can be run on CPU only. Furthermore, $k$NN-LM use \textit{FAISS}, which is an efficient library that allows them to quickly find $k$NNs. 

One interesting detail to note in \citep{khandelwal20generalization} which was crucial for this work, was that the authors tried both the inner product and the L2 for their distance metric in $k$NN. They concluded that L2 worked significantly better. This observation implies the fact that the default training recipe of LMs \textit{implicitly} prefers one distance over another. We then ask the question whether one could train in such a way that we could adapt the model to the post-hoc $k$NN with a given distance metric. In the next section, we describe how to adapt the training of the LM, with $k$NN and the corresponding L2 distance metric in mind.

\section{Proposed Method}
In our initial attempt, we experimented with the idea of explicitly minimizing the L2 distance between context vectors which predict the same target word. This strategy directly mirrors the use of context vectors at the $k$NN step, and we hope that training the representations in a way similar to testing will further improve the effectiveness of $k$NN LM. However, a naive implementation of it is infeasible. We then resorted to a MOCO \citep{he2020momentum} style training scheme. Specifically, for each target word $w$, we construct a queue $Q$ of fixed length $L$, which stores the recent $L$ context representations for $w$. During training, we optimize a regularized objective as follows:

\begin{align}
    \label{moco}
    \mathcal{L}_{new} = L_{CE} + \omega \sum_{j=1}^{\texttt{N}}\sum_{i=1}^{\texttt{L}} ||sg( Q^{w_j}_i) - r_j ||^2,
\end{align}
where $N$ is the batch size, $r_j$ is the context representation of the $j$th word, $Q^{w_j}$ is the queue corresponding to the $j$th target word $w_j$; $\omega$ is the regularization parameter; $sg(\cdot)$ is the stop gradient operator. Specifically, $Q$ is updated with a momentum target encoding network which is initialized with the same parameters of the LM, similar to MOCO \citep{he2020momentum}. 

\begin{figure}[!htp]
    \centering
    \includegraphics[width=0.45\textwidth]{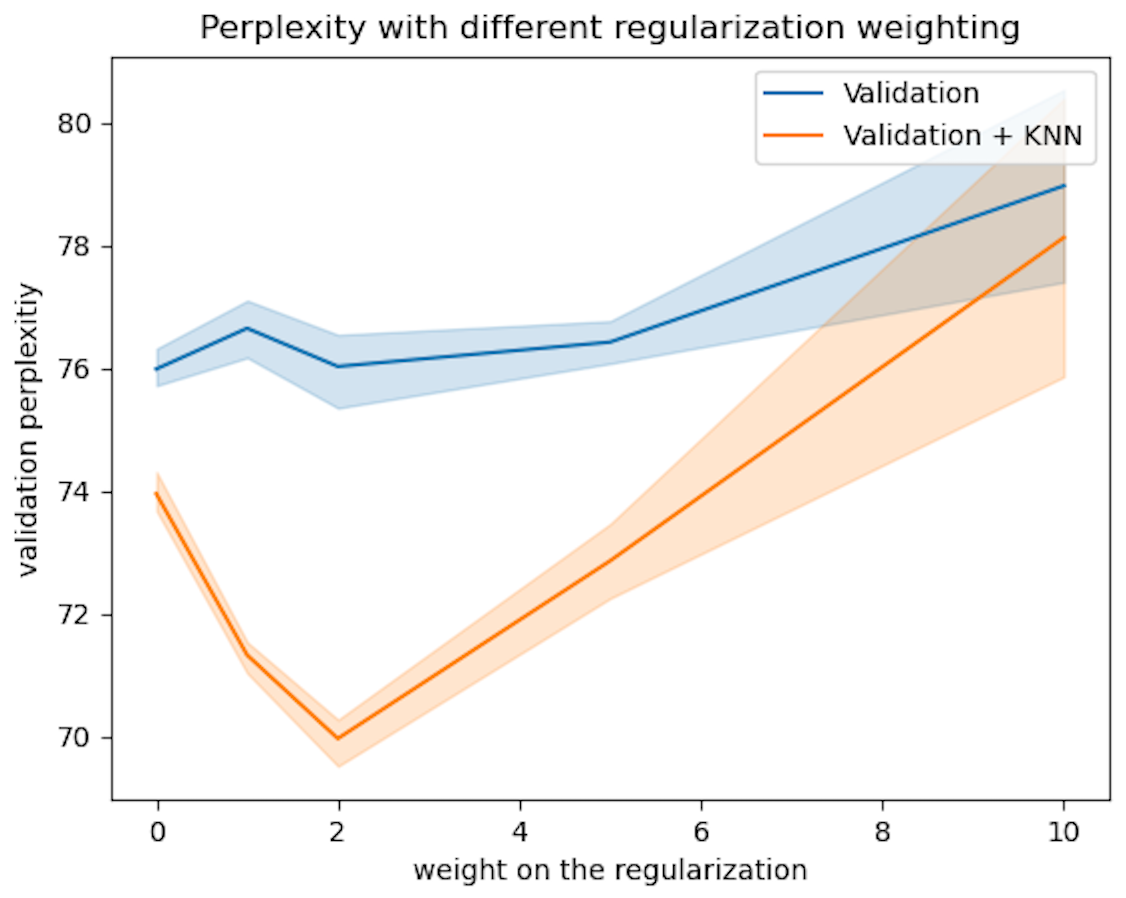}
    \caption{Validation perplexity on \texttt{WIKI-2} of the LM before (blue) and after (orange) adding $k$NN. NOTE: \texttt{weight=0} corresponds to the standard version that does not include our added MOCO-style regularization term i.e. original $k$NN-LM  from \cite{khandelwal20generalization}}
        \label{fig:wiki2_res}
\end{figure}

Empirically, we found that Equation \ref{moco} provides a practical solution and yields improved representations for the $k$nn LM, as shown in Fig \ref{fig:wiki2_res}. However, the use of the queue and momentum target network still adds overhead to a large scale model training. Hence we tried to decrease $Q$ and $L$, which interestingly did not decrease the performance at all and therefore, to promote efficiency, we tested an even simpler formulation, where we replace $Q$ with all \textbf{zero} vectors. This eliminates the need of explicitly constructing and updating the queue, while instead encourages the model to learning conservative representations w.r.t. the L2 norms of its context representations. The corresponding loss is as follows:
\begin{align}
\label{newloss}
\mathcal{L}_{new} = L_{CE} + \omega \sum_{j=1}^{\texttt{N}} ||r_j ||^2.
\end{align}
To our surprise, Equation \ref{newloss} yields similar performance to Equation \ref{moco} in practice, while being much easier to implement and tune. This is a new finding which we will try to explain in the ablation study. We thus use Equation \ref{newloss} as the default loss function in our experiments unless otherwise mentioned.

\section{Experiments}
We tested our method on the \texttt{WIKI-2} and \texttt{WIKI-103} datasets, which are widely used benchmarks for language modelling.  We are interested in demonstrating two empirical results: improved performance using our approach over that of $k$NN-LM, and exploring a possible mechanism for this improved performance.

\begin{table*}[!h]
  \centering
  \scalebox{0.9}{
  \begin{tabular}{lcccc}
    \toprule
         & $k$NN-LM  $(\omega\!=\!0)$    &  $\omega\!=\!0.1$ & $\omega\!=\!1.0$ & $\omega\!=\!10.0$ \\
    \midrule
    Train Ppl. LM & 19.99  & 20.05 & 20.11 & 21.37     \\
    Valid Ppl. LM & 75.96  & 75.68 & 76.37 & 81,29     \\
    Valid Ppl. $k$NN-LM & 74.11  & 73.13 & \textbf{70.63} & 80.52     \\
    \bottomrule
  \end{tabular}}
    \caption{Experiments on \texttt{WIKI-2} with corresponding validation perplexity using L2 regularization.}
    \label{tab:wiki2}
\end{table*}

\subsection{Experimental setup}
\textbf{Dataset}

\texttt{WIKI-2} is a benchmark with 30k word vocabulary and consists of 2M tokens. \texttt{WIKI-103} is a benchmark with 250k word vocabulary and consisting of 103M tokens.

\textbf{Language Model Architecture}
For the language model architecture, we will be using the exact setup as described in \citep{khandelwal20generalization}. This setup consists of the language model by \citep{baevski2018adaptive}, which consists of 16 layers, each with 16 self-attention heads, 1024 dimensional hidden states, and 4096 dimensional feedforward layers. Thus, following \citep{baevski2018adaptive}, this LM has adaptive inputs and an adaptive softmax \citep{joulin2017efficient} with tied weights \citep{press2016using} for all our experiments. We trained the each language model on a \texttt{Tesla V100} with \texttt{400GB} if RAM. 

In addition, we follow the exact same training procedure as in \cite{khandelwal20generalization} and refer to their paper for further details on the training parameters. The only difference in terms of implementation is the MOCO style learner as well as the L2 regularization added to the final layer. Lastly, we would like to note that while crossvalidating though the interpolation parameter $\lambda$ we note that for all models, $\lambda=0.3$ works the best which is in accordance to the finding in \cite{khandelwal20generalization}.

\subsection{Experiments on \texttt{WIKI-2}}

We first apply our proposed method on the standard \texttt{WIKI-2} dataset, where we run each configuration 5 times and plot the standard deviation, as seen in Fig. \ref{fig:wiki2_res}. Note that $\omega=0$ in Fig. \ref{fig:wiki2_res} corresponds to the standard $k$NN-LM version, i.e. without the added term in the loss. Comparing Figure~\ref{fig:wiki2_res} and Table~\ref{tab:wiki2} to see that the MOCO and L2 approaches produces similar results. From these results, we note the following phenomena:

\begin{enumerate}
    \item A clear "U"-shape demonstrating the added benefit of our loss term on the validation perplexity of the LM for moderate values of $\omega$. 
    \item Training performance does not decrease for moderate values of $\omega$, showing that the extra term does not destroy training and generalization of the standard LM.
    \item There is no difference in terms of validation perplexity between the standard LM and our version \textbf{before} applying $k$NN, but there is a significant difference \textbf{after} applying $k$NN.  Our approach likely finds a different local minimum for the language model that is better suited for $k$NN.
\end{enumerate}

\begin{table*}[!h]
  \centering
  \scalebox{0.9}{
  \begin{tabular}{llll}
    \toprule
         & $k$NN-LM  $(\omega\!=\!0)$    & $k$NN-LM  $(\omega\!=\!1)$ & $k$NN-LM  $(\omega\!=\!10)$ \\
    \midrule
    Train Ppl. LM & 11.31  & 11.24 & \textbf{11.07}     \\
    Valid Ppl. LM & 18.00  & 17.95 & \textbf{17.71}     \\
    Valid Ppl. $k$NN-LM & 16.09  & \textbf{15.89} & 17.46     \\
    \bottomrule
  \end{tabular}}
    \caption{Experiments on \texttt{WIKI-103}}
    \label{tab:wiki103}
\end{table*}
\subsection{Experiments on \texttt{WIKI-103}}
We illustrate our findings on the more challenging \texttt{WIKI-103} dataset and demonstrate that our L2 fix significantly improves the performance of the LM. In the Table~\ref{tab:wiki103}, we illustrate that when changing the regularization strength we again see a significant gain in performance when adding our regularization during training of the LM. Due to the computational costs when training these models, we resort to the same hyperparameters as in the \texttt{WIKI-2} dataset and hence present fair comparisons of the different variants of the model.


Note that again, we see significant improvements in terms of validation perplexity when using the $k$NN-LM scheme by simply adding a L2 regularization when training the language model.

On another note, when taking a closer look at the validation perplexity before applying $k$NN, we note that $\omega=10$ seems to have lowest validation perplexity. This better generalization phenomenon is interesting and has recently been noted in the machine vision community in the context of investigating the regularization effects of batch normalization in classification settings \citep{dauphindeconstructing}. This also relates to the findings in \citep{merity2017regularizing}, who used L2 regularization in LSTMs. In this paper, we have found initial indications that the L2 regularization on the activations might also be useful for Transformer models. 

Finally, we believe that these two standard benchmark datasets in language modelling are sufficient evidence to demonstrate the merit our of findings.


\vspace{-0.3cm}
\subsection{Further investigations into the representations and possible explanations}

\begin{figure*}[!htp]
    \centering
    \includegraphics[width=0.46\textwidth]{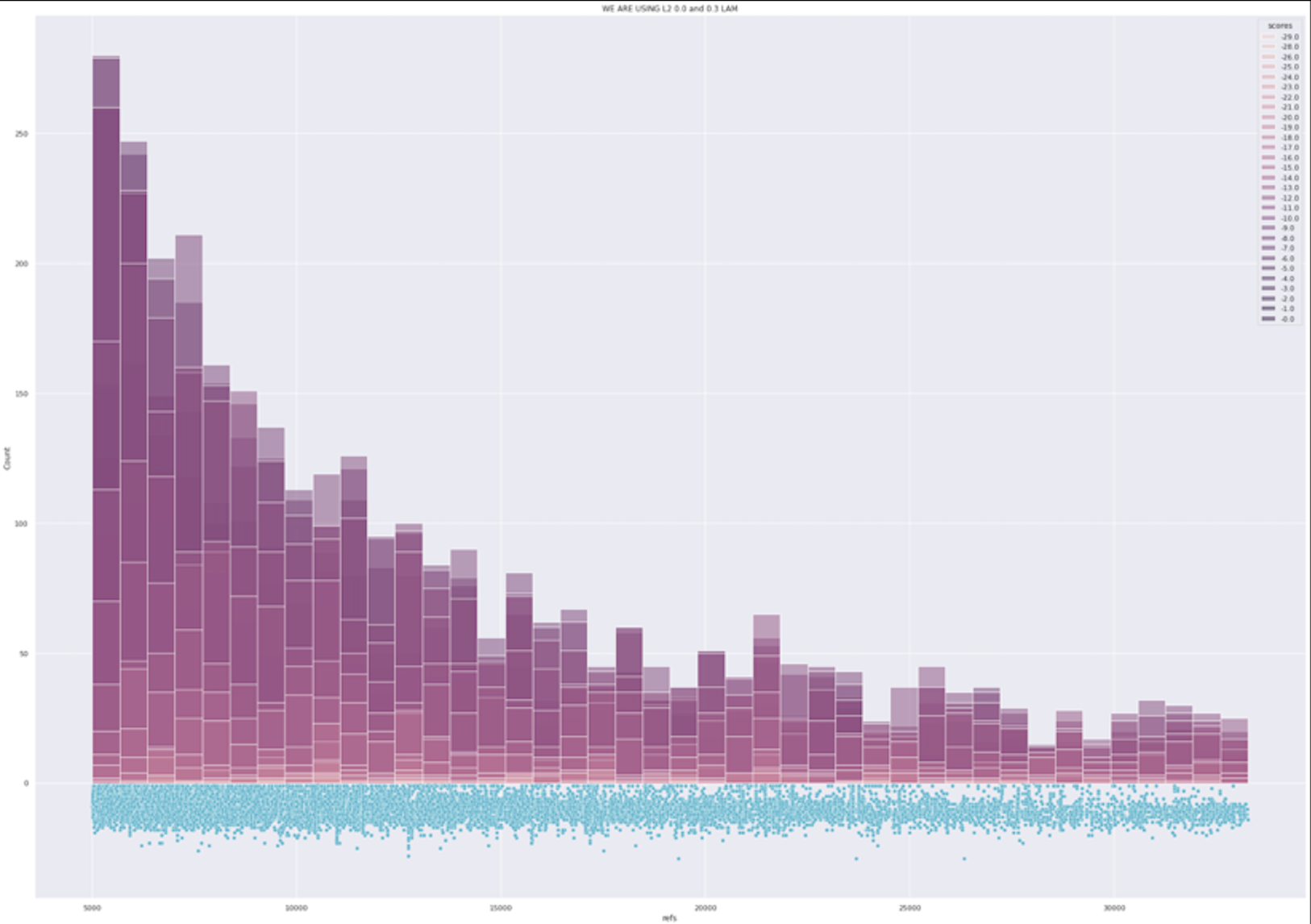}
    \includegraphics[width=0.46\textwidth]{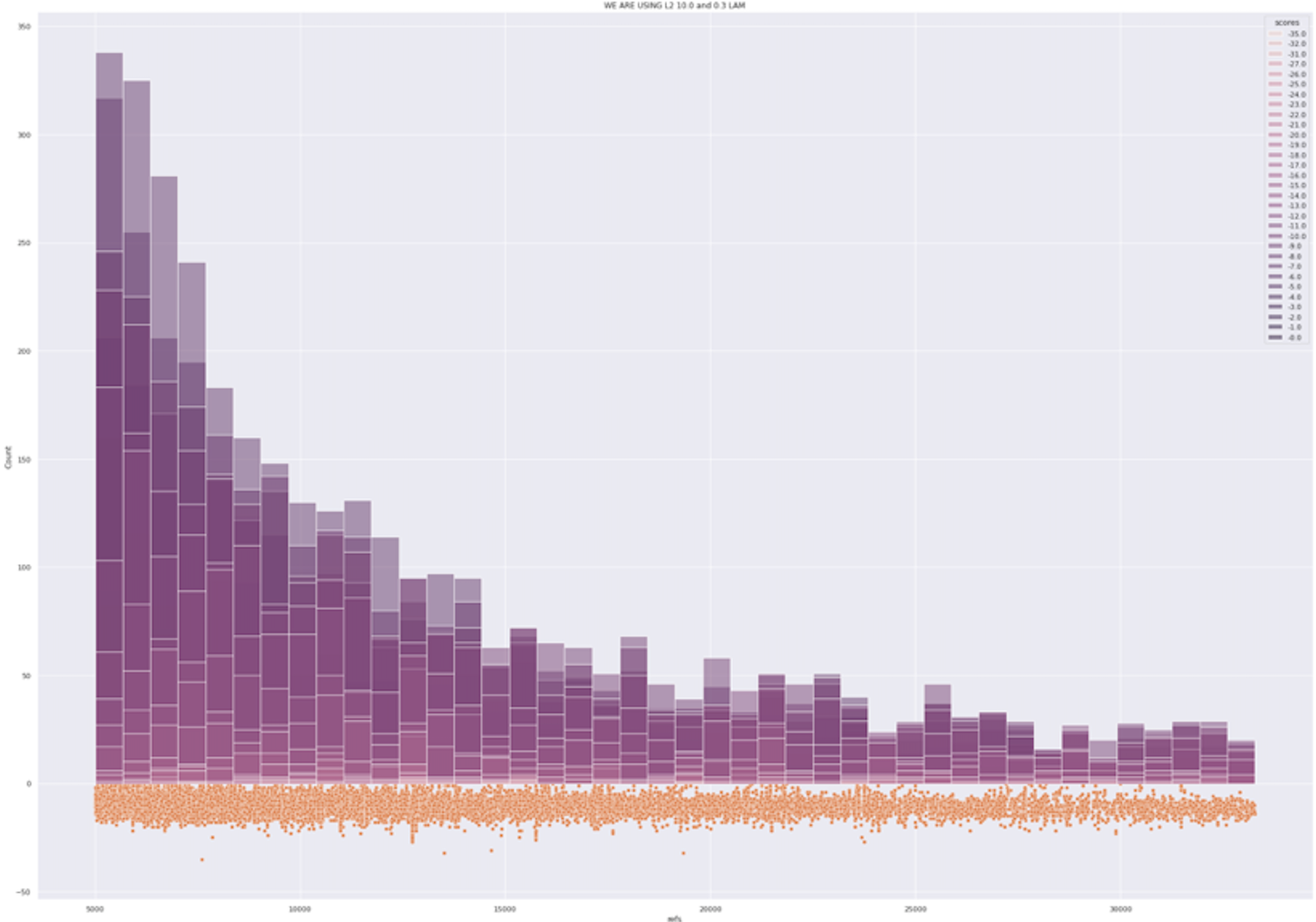}
    \caption{Frequency/Loss histograms. The x-axis denotes the frequency of the word with high-frequency words to the left.  The y-axis denotes the number of words with $x$ frequency and the colors of each bar represents the loss accumulated.  (LEFT) Standard LM after $k$NN, (RIGHT) Our L2 regularized LM after $k$NN.  }
    \label{fig:my_label}
\end{figure*}

We first examine the effect that target word frequency has on the loss. Figure~\ref{fig:my_label} shows a histogram of word frequency, where the color represents the respective losses each word incurred.

Note that firstly, there is little difference in loss for the less frequent words on the right side of the graph. If we shift our attention to the more frequent words however, we see a different picture. Looking at our L2 regularized model, we note that for the most frequent words, our model seems to incur lower loss (see the brighter colors bars at the peak of the histograms) compared to the standard LM with $k$NN. This observation suggests that the main differences in terms of representations come from the frequent words rather than rare ones. 


Secondly, knowing that the main differences are within the words that are most frequent, we investigated these representations in more detail. In particular, we analyzed the most frequent words and divided the data into "\textit{high scores}" meaning they contributed a lot to the loss (bad predictions) and "\textit{low scores}" meaning they are good predictions i.e. they contributed a little to the loss.

\begin{figure*}[!htp]
     \centering
     \includegraphics[width=0.48\textwidth]{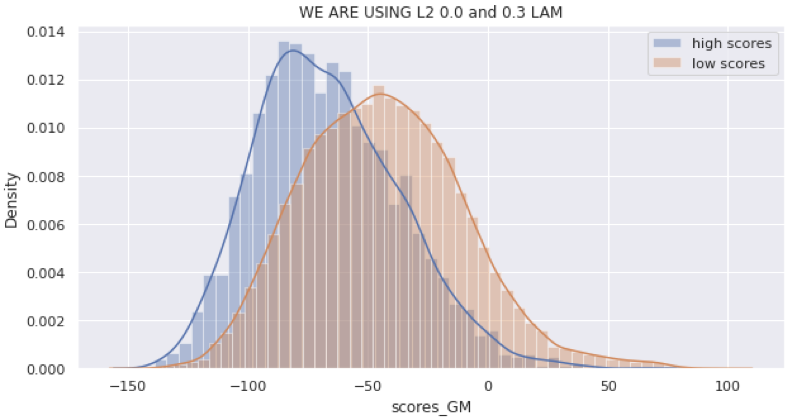}
    \includegraphics[width=0.48\textwidth]{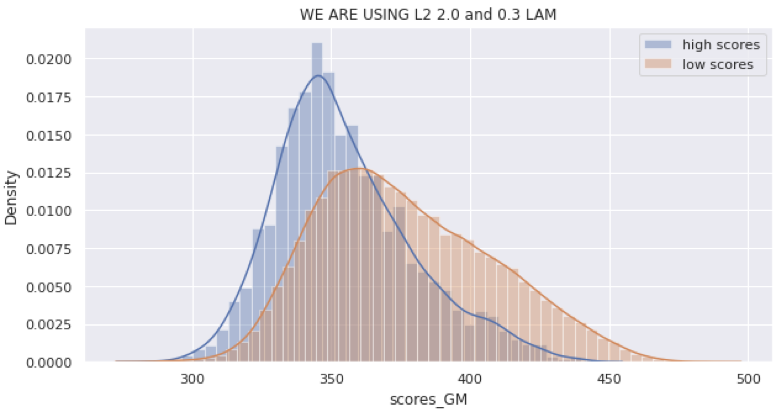}
    \caption{$x$-axis denotes the Loglikelihood under the Gaussian Mixture (GM). $y$-axis denotes the normalized histogram. (TOP) Standard training of then LM (BOTTOM) using a L2 regularization to train the LM.}
    \label{fig:hists}
\end{figure*}

We employed a simple mixture of Gaussians model (GMM) ($m=10$) and used the log-likelihood as an indicator for how well the data are clustered. GMMs allows us to put probability mass on each of the representations and given that we are using a mixture of gaussians, we inherently captures clusters. GMMs can be thought of as "putting multiple gaussian distributions on the data". Intuitively, this means that if the likelihood of the GMM is high, the representations can be easily captured using a mixture of gaussians, which is indicative of being more clustered i.e. close to one of the gaussian mixture means.

In Figure~\ref{fig:hists} we compare the distributions of the loglikelihoods for the representations that have been trained using the standard LM and our modified L2 regularization. In particular, for each representations we obtain the corresponding likelihood from the GMM ($x$-axis on Figure~\ref{fig:hists}). As mentioned before, we split the words into "\textit{high scores}" and "\textit{low scores}" and plot their histograms in blue and orange respectively. Fig. \ref{fig:hists} demonstrates that when using a L2 regularization (right), the representations that contribute less to the loss (shown in orange) are better fitted by the GMM than when training the model in the standard way. This can be seen by the increase in likelihood, i.e. shift to the right on the $x$-axis. What is interesting is that the difference between the likelihoods of the "\textit{high scores}" and "\textit{low scores}" varies much more dramatically in the L2 regularized case.  Hence, one of our hypotheses is that $k$NN-LM improves the classification accuracy mostly for the non-frequent words \citep{khandelwal20generalization}, whereas our proposed method with L2 regularization, in addition, also improves the classification accuracy of the frequent words by clustering them closer together and hence allowing $k$NN-LM to work better.

\section{Conclusion}
\vspace{-0.3cm}
In conclusion, we propose a useful training mechanism that is inspired by the fact that the post-hoc application of $k$NN seems to significantly improve the performance of standard LMs. We have found that training a LM with L2 regularization at the final layer, i.e. layer which is used for the post-hoc $k$NN search, improves validation performance. We have also found initial indications that the L2 regularization mostly improves performance for the most frequent, lower-loss words. 

These findings motivate exploring L2 regularization in different Transformer architectures and LM tasks. Furthermore, using training data at inference time is an interesting direction for improving model performance, and we hope that our approach encourages future work in this area.
\newpage

\bibliography{anthology,custom}
\bibliographystyle{acl_natbib}

\end{document}